\documentclass{article}

\usepackage{amsmath,amsfonts}
\usepackage{algorithmic}
\usepackage{graphicx,color}
\usepackage{caption}
\usepackage[numbers]{natbib}
\usepackage{hyperref}
\usepackage{natbib}
\usepackage{authblk}

\usepackage{tikz}
\usetikzlibrary{shapes,arrows,shadows,patterns,positioning,trees,positioning,decorations.pathreplacing}
\usetikzlibrary{arrows,shapes,positioning,shadows,trees}
\tikzset{
  basic/.style  = {draw, text width=2cm, drop shadow, font=\sffamily, rectangle},
  root/.style   = {basic, rounded corners=2pt, thin, align=center,fill=blue!30,text width=12em},
  level 2/.style = {basic, rounded corners=7pt, thin, align=center, fill=orange!70, text width=5em, yshift=-55pt},
  level 3/.style = {basic, thin, align=left, fill=yellow!60, text width=6.5em}
}

\AtBeginDocument{%
  }

\begin{document}

\title{LLM-based event log analysis techniques: A survey}

\author[$\dagger$]{Siraaj Akhtar}
\author[$\dagger$]{Saad Khan}
\author[$\dagger$]{ Simon Parkinson}
\affil[$\dagger$]{Department of Computer Science, University of Huddersfield, UK}
\date{}

\maketitle


\begin{abstract}
Event log analysis is an important task that security professionals undertake. Event logs record key information on activities that occur on computing devices, and due to the substantial number of events generated, they consume a large amount of time and resources to analyse. This demanding and repetitive task is also prone to errors. To address these concerns, researchers have developed automated techniques to improve the event log analysis process. Large Language Models (LLMs) have recently demonstrated the ability to successfully perform a wide range of tasks that individuals would usually partake in, to high standards, and at a pace and degree of complexity that outperform humans. Due to this, researchers are rapidly investigating the use of LLMs for event log analysis. This includes fine-tuning, Retrieval-Augmented Generation (RAG) and in-context learning, which affect performance. These works demonstrate good progress, yet there is a need to understand the developing body of knowledge, identify commonalities between works, and identify key challenges and potential solutions to further developments in this domain. This paper aims to survey LLM-based event log analysis techniques, providing readers with an in-depth overview of the domain, gaps identified in previous research, and concluding with potential avenues to explore in future.
\end{abstract}




\maketitle

\section{Introduction}
This paper provides an overview of developments, up to the time of writing, in the research field of using Large Language Models (LLMs) to perform event log analysis. Event logging is used to record events that occur in various systems and are of varying types, including system, application, and network logs~\cite{kurniati2016implementing}. Performing log analysis is necessary as it provides insight into how systems are performing and threats to optimal performance. If issues are found, then security professionals can address them. However, because of the significant number of events being generated, they require a large amount of time to analyse. In addition, since security professionals perform the task regularly, they can suffer fatigue~\cite{ban2021combat} and as a result, logs that may provide important insights could go undetected. Due to these concerns, research needs to be conducted to improve the efficiency of event log analysis.

Over the past few years, LLMs have gained attention in various fields due to their ability to perform tasks that humans usually undertake accurately, and at times exceed the performance that humans can achieve~\cite{yang2024harnessing}. It has been highlighted that, specifically for data analysis tasks, they can perform at a high level~\cite{nejjar2023llms}. As data analysis tasks are repetitive and the use of humans to perform these tasks is seen to be inefficient, deploying LLMs in these scenarios is a positive development.

\begin{figure*}
\centering
\large
\begin{tikzpicture}[
  level 1/.style={sibling distance=37mm},
  edge from parent/.style={->,draw},
  >=latex,scale=0.63,transform shape,auto]
\node[root] {LLM-based Event Log Analysis}
    child {node[level 2] (c1) {Anomaly Detection}}
    child {node[level 2] (c2) {Fault Monitoring}}
    child {node[level 2] (c3) {Log Parsing}}
    child {node[level 2] (c4) {Root Cause Analysis}}
    child {node[level 2] (c5) {SIEM}}
    child {node[level 2] (c6) {System Health Monitoring}}
    child {node[level 2] (c7) {Threat Detection}};

\begin{scope}[every node/.style={level 3}]

\node [below of = c1, xshift=15pt, yshift=-20pt] (c11) {Exploratory Research~\cite{kethireddyai,srivatsalogsense}};
\node [below of = c11, yshift=-15pt] (c12) {BERT~\cite{chen2022bert,corbelle2024semantic,cometti2024real,sun4978351improving}};
\node [below of = c12, yshift=-10pt] (c13) {GPT Series~\cite{qi2023loggpt,ali2023huntgpt,hadadi2024anomaly}};
\node [below of = c13, yshift=-10pt] (c14) {Comparison~\cite{balasubramanian2023transformer,mannam2023optimizing}};
\node [below of = c14, yshift=-22pt] (c15) {Machine Learning Approaches~\cite{10589612,li2024ids,fariha2024log}};
\node [below of = c15, yshift=-28pt] (c16) {In-Context Learning~\cite{egersdoerfer2023early,jin2024large}};
\node [below of = c16, yshift=-20pt] (c17) {Differing Approaches~\cite{zheng2023logdapt,10414427,guo2024logformer,liao2024detecting,10707565}};
\node [below of = c17, yshift=-28pt] (c18) {Reducing Resource Consumption~\cite{song2024audit,yang2024anomaly}};

\node [below of = c2, xshift=15pt, yshift=-15pt] (c21) {Exploratory Research~\cite{mudgal2024crasheventllm}};
\node [below of = c21, yshift=-10pt] (c22) {Crash Prediction~\cite{huang2024demystifying}};
\node [below of = c22, yshift=-25pt] (c23) {Fault-Indicating Information Extraction~\cite{shan2024face}};

\node [below of = c3, xshift=15pt, yshift=-15pt] (c31) {Exploratory Studies~\cite{le2023log,ma2024llmparser}};
\node [below of = c31, yshift=-15pt] (c32) {Use of Different LLMs~\cite{xu2024divlog,zhou2024leveraging,astekin2024comparative}};
\node [below of = c32, yshift=-28pt] (c33) {In-Context Learning~\cite{zhi2024llm,yu2024loggenius,jiang2024lilaclogparsingusing,wu2024log,vaarandi2024using,zhong2024logparser,huang2024ulog,ma2024librelog}};
\node [below of = c33, yshift=-22pt]  (c34) {Fine-Tuning~\cite{chen2024high}};

\node [below of = c4, xshift=15pt, yshift=-15pt]  (c41) {Exploratory Studies~\cite{ahmed2023recommending}};
\node [below of = c41, yshift=-15pt] (c42) {Few-Shot Learning~\cite{10.1145/3663529.3663841,chen2024automatic,zhang2024automated}};
\node [below of = c42, yshift=-15pt] (c43) {Fine-Tuning~\cite{mandakath2024root}};

\node [below of = c5, xshift=17pt, yshift=-20pt] (c51) {Exploratory Studies~\cite{sultana2023towards,kaheh2023cyber,yao20235g}};
\node [below of = c51, yshift=-30pt] (c52) {Evaluating Models Over Time~\cite{ahmed2024prompting,karlsen2024benchmarking,jonkhout2024evaluating,hermann2024gpt,oniagbi2024evaluation}};

\node [below of = c6, xshift=15pt, yshift=-20pt]  (c61) {No Specific Research Found};

\node [below of = c7, xshift=15pt, yshift=-20pt] (c71) {Use Cases~\cite{chen2024survey,khanllms,patil2024leveraging}};
\node [below of = c71, yshift=-22pt] (c72) {Machine Learning Approaches~\cite{ferrag2024revolutionizing,hasan2024distributed}};
\node [below of = c72, yshift=-23pt] (c73) {Comparison\\~\cite{sanchez2024transfer,houssel2024towards}};
\node [below of = c73, yshift=-15pt] (c74) {Fine-tuning~\cite{novaklightweight,lanka2024intelligent,rahmani2024integrating}};
\node [below of = c74, yshift=-15pt] (c75) {Concept Drift~\cite{sundin2024ai}};

\end{scope}

\foreach \value in {1,...,8}
  \draw[->] (c1.195) |- (c1\value.west);

\foreach \value in {1,...,3}
  \draw[->] (c2.195) |- (c2\value.west);

\foreach \value in {1,...,4}
  \draw[->] (c3.195) |- (c3\value.west);
  
\foreach \value in {1,...,3}
  \draw[->] (c4.195) |- (c4\value.west);

\foreach \value in {1,...,2}
  \draw[->] (c5.195) |- (c5\value.west);
  
  \draw[->] (c6.195) |- (c61.west);

\foreach \value in {1,...,5}
  \draw[->] (c7.195) |- (c7\value.west);

\end{tikzpicture}
\caption{Taxonomy Diagram detailing event log analysis tasks and subtasks found in the literature review~\cite{10.1145/3460345, 6261962, li2017flap, 10.1145/3400286.3418261}}
\label{fig:TaxonomyDiagram}
\end{figure*}
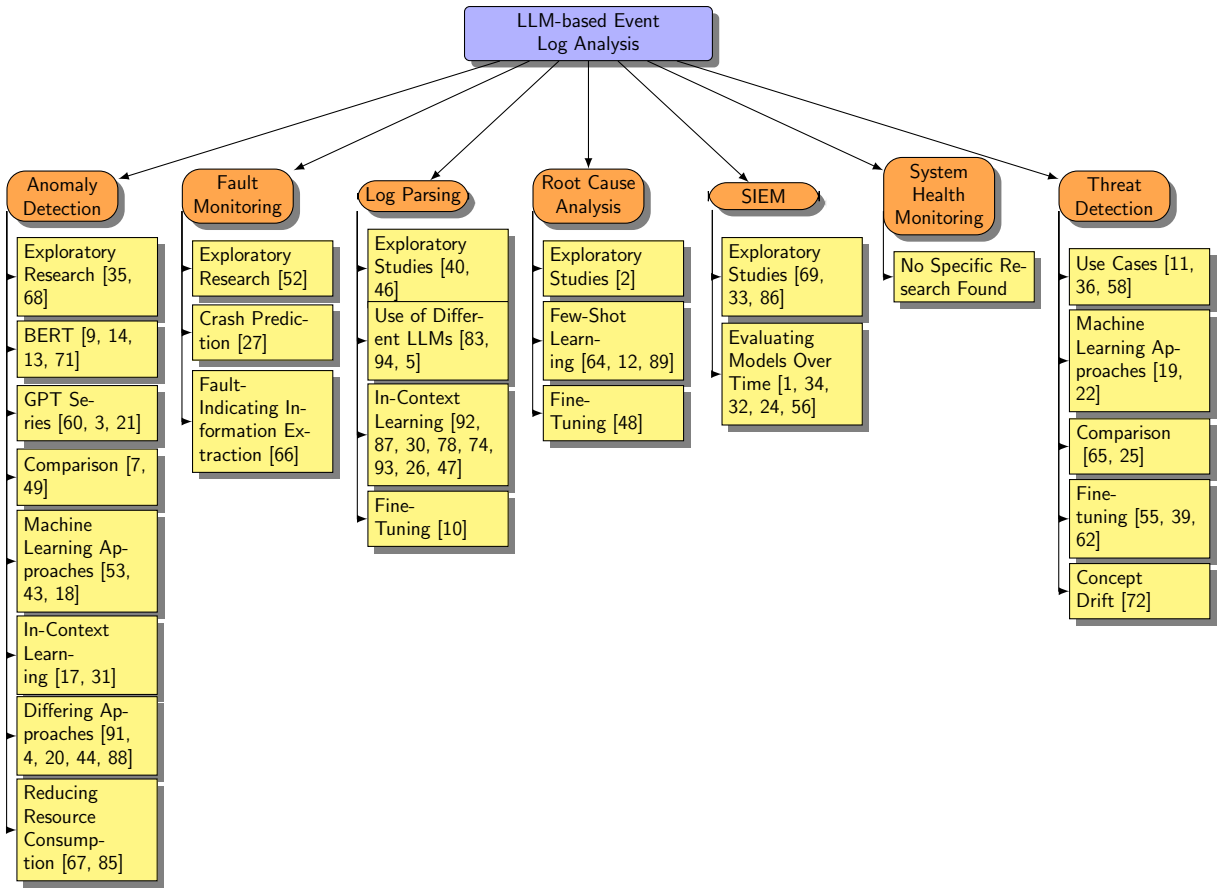

Following on from early developments, researchers saw the potential of using LLMs to perform event log analysis tasks. At the beginning stages, researchers investigated how well different models could perform event log analysis, researchers deduced that although closed-source LLMs such as the GPT series and Claude~\cite{balasubramanian2023transformer} performed event log analysis slightly better when compared to their open-source counterparts, they were not suitable to be used as they require data to be sent to the developer's servers, which is not desirable for most organisations who have reservations over releasing data that could be used to identify security weaknesses. Following this, researchers investigated how different factors affect event log analysis so that developers can use those shown to improve performance and abstain from those that decrease it. These factors include in-context learning, fine-tuning, and RAG~\cite{egersdoerfer2023early}. 

To continue to develop, it is essential to grasp the progress made within the field and understand the challenges and limitations. This will produce a clear picture of where to focus our attention in future work, and ensure that the work researchers undertake is meaningful and furthers development. To address this, with this article we intend to provide an overview of all research, to the best of our knowledge, up to this current point, within the field.

Following this section, we begin by illustrating the various components that form our methodology, used to obtain all research articles analysed within this paper. Preceding, we undertake a systematic literature review, analysing different topics relevant to the use of LLMs for event log analysis. This results in the discovery of various gaps and limitations which can be investigated in future work. We provide an analysis of these gaps, highlighting what they are and potential solutions that future researchers could use to address them. Finally, we conclude with an overview of the directions that researchers could take on the basis of this article. Figure~\ref{fig:TaxonomyDiagram} provides an overview of the groupings of works identified in the examined literature. The figure is provided as an illustrative aid to the reader.

\section{Methodology}

The methodology of this paper was, to the best of our ability, to gather all articles based on the use of LLMs for event log analysis. Google Scholar was used to search for articles hosted in the following research repositories, such as Elsevier, ScienceDirect, Springer, IEEE Xplore, etc. After the articles were recovered, we analysed their titles and abstracts to ensure that they were relevant, and those that were not were removed. 

\begin{figure*}
\centering
\large
\begin{tikzpicture}[node distance=4cm, every node/.style={draw, text width=3cm, fill=yellow!50, align=center, minimum height=1.5cm}]
    \node (box1) {Articles obtained from research databases (94)};
    \node (box2) [right of=box1] {After duplicates discarded (84)};
    \node (box3) [right of=box2] {Abstract read to determine the relevancy to topic};
    \node (box4) [right of=box3] {Articles reviewed};

    \draw[->] (box1) -- (box2);
    \draw[->] (box2) -- (box3);
    \draw[->] (box3) -- (box4);
\end{tikzpicture}

\caption{Process of PRISMA framework to find relevant and review relevant articles.}
\label{fig:TaxonomyDiagram}
\end{figure*}
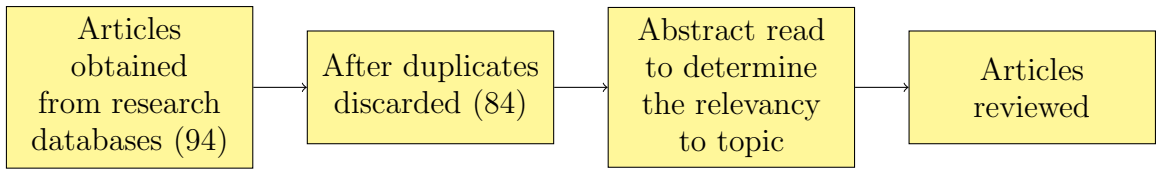


In this project, we have used the PRISMA framework to gather relevant articles, as shown in Figure~\ref{fig:prisma}. Each step/phase of the framework is explained in the following.

\subsubsection{Obtaining Articles}
To obtain all articles published on our topic, all topic-specific search terms would need to be used. The following search terms were used: ``system logs llm'', ``large language model event logs'', and ``event log llm''. In addition to the full and abbreviated forms of words, as an example, ``LLMs'' instead of ``Large Language Models'', and changing between the plural and singular versions of words. The collected articles are published in conferences and journal articles.

\subsubsection{Discarding Duplicates}
After obtaining the articles, 10 were found to be the same article, because they were present in multiple repositories. These articles were subsequently discarded, as it is unnecessary to analyse an article more than once.

\subsubsection{Discarding Irrelevant Articles}
After reading the abstract of the remaining articles, it was found that some of the articles collected, although their titles suggested relevance, were not relevant. This was either because they focussed solely upon LLMs, and event log analysis was restricted to a minuscule mention, due to the facts that they were not used to perform event log analysis tasks specifically, or they were specific to event log analysis tasks and not focussing on the use of LLMs to optimise these tasks. As a result, they were removed from our study to ensure that our study stays within its established boundaries.

\subsubsection{Analysis}
Finally, we were left with 75 articles (10 which can be found in the literature review section of this report) that were different articles and relevant to the field. Our analysis used the following format: explaining the objectives the authors intended to address, the methodology that they used to achieve these objectives, results (which mainly consisted of comparisons to other methods to highlight if the solution was an improvement) and a conclusion which summarised limitations to the studies and future work that can be conducted based on the article in its entirety. The various articles (within the categories mentioned previously) were organised into paragraphs, based on the year published with the techniques and methods employed. An example is that articles using fine-tuning together, compared to those using in-context learning, were in separate paragraphs.

\section{Literature Review}
\subsection{Anomaly Detection}
\subsubsection{Early Research}
A large majority of the literature related to the topic of using LLMs for event log analysis focusses on anomaly detection. First, we will look at~\cite{kethireddyai} who present an insider detection system which includes event logs as well as other data, to create a picture of the usual behaviour patterns of users. To begin, sensitive and user-identifiable information is removed to ensure privacy, it is then passed into the LLM to understand the patterns, normal behaviour patterns are established by using clustering, PCA, and auto-encoders, and finally the anomaly is detected using machine learning algorithms and NLP methods. The results show that there was an increase in the accuracy, precision, recall and F1 scores and the false positives were reduced by 40\%. A limitation of this study is that it does not mention which LLM model was used. As a result of this, future work can investigate which LLMs can perform event log analysis, and in particular anomaly detection, accurately and efficiently. Similarly,~\cite{srivatsalogsense} introduces LogSense, which uses both LogGPT and RAPID for anomaly detection. RAPID analyses individual event logs, and LogGPT analyses event logs as a whole. In addition, at the end of each day, the model is fine-tuned to increase accuracy, as event logs change over time. However, a limitation is that there is no data presented to show how well the model performs and no comparison with other models, so we are unable to identify its performance on anomaly detection in event logs. Future work based on this model can address the high false positive rate, as the models are developed to reduce the time security professionals spend analysing data, but the effects of this benefit are reduced with the presence of this limitation.

\subsubsection{BERT}
Moving on, we will look at models and solutions which are based on the BERT architecture.~\cite{chen2022bert} introduce BERT-Log. Firstly, it structures the raw event logs and puts them into groups based on the order in which they occurred. The authors mention that their work is ``the first to utilize node ID and time to form log sequence''. After this, the logs are encoded in token form and put into vectors and then anomaly detection takes place. The authors tested BERT-Log on the BGL and HDFS datasets, changing the amount of data used for training, and then compared it to other methods. The results show that BERT-Log consistently received F1-Scores of 0.99 on both datasets, whereas the other models performed better on the HDFS dataset and differed in performance depending on the amount of data used for training. This shows that BERT-Log performs consistently when the amount of data used for training changes, and without a noticeable decrease in performance, when trained on more complex datasets such as BGL. The authors mention that future work based on this article can look at improving training efficiency and as a result reducing resource consumption. 

Similarly,~\cite{corbelle2024semantic} introduces a BERT-based solution called BERT20M to produce a model for anomaly detection, reducing the size of the data. They propose using hierarchical classification, where log data is grouped into themes or categories. They conducted two tests; the results from the first test show that there is a significant decrease in the amount of time to process data, total number of lines, and count of tokens, for both 2k and 20mg, when hierarchical classification was applied, compared to the original data. For the second test, their model had the second lowest precision, recall, and F1 scores. As a result of this, a suggestion for future research is to research other methods of reducing the size of data, which can produce results which are the same or better than existing solutions. A different approach is taken by~\cite{cometti2024real} who use the RAPID method to find anomalies in event logs. Their method involved fine-tuning a DistilBERT model on a dataset comprised of event logs, obtained from the industry, and then comparing it to the base DistilBERT model. The test produced surprising results, as both the base and the fine-tuned models had F1 scores of 0.94, Although this score looks positive, they highlight that there are a large number of false positives (normal logs classed as anomalies). A limitation of this method is that, although no training is required, specifically for anomaly detection, fine-tuning still needs to take place and this would still consume a large amount of resources. Future work based on this article can look into reducing the false positive rate. 

A similar approach is also taken by~\cite{sun4978351improving} introducing LogRoBERTa, their paper and method mention that previous work produced high F1 scores, but as security professionals need anomalies to be identified quickly, they attempt to reduce the time taken for anomaly detection. Specifically, they evaluate removing the parsing step as they suggest that the models can understand raw event logs without the need for parsing. RoBERTa is selected as it is an improvement of the BERT model because certain aspects such as layers and attention heads are increased, resulting in an improvement in understanding. Their results showed that LogRoBERTa outperformed the other methods in both data sets, highlighting the improvement of RoBERTa compared to BERT in anomaly detection. In addition, LogRoBERTa's F1 scores did not significantly change when using a parser, and the same was found with LogBERT. However, this was only based on a test using one dataset, with three parsers and with methods based on BERT. Thus, a suggestion for future work is to verify the results and suggestions in this paper by evaluating using other LLMs, datasets and parsers to prove that the use of parsers does not significantly improve anomaly detection with LLM models.

\subsubsection{GPT Series}
We will now look into research published on the GPT series. Beginning,~\cite{qi2023loggpt} presents LogGPT which was one of the first models to be used for anomaly detection using LLMs. Their method specifically uses ChatGPT and begins by parsing the logs to a suitable format, implementing few-shot learning, showing the model how to perform the task, and a ``response parser'' where the output is parsed and then outputted to the user. The methodology of testing was to compare its performance using the BGL and Spirit datasets against three machine learning models with both the model using few-shot and zero-shot learning. The results showed that LogGPT performed the second best on the BGL dataset and achieved comparable results, to two of the three machine learning models, but at times it achieved low scores on the Spirit dataset. In addition, generally, LogGPT performed better with few-shot learning than with zero-shot. As this was an early piece of work, suggestions based on the method would be to use methods such as fine-tuning or RAG to improve the results, in addition, to researching the use of offline models. 

Another solution based on the GPT architecture is presented by~\cite{ali2023huntgpt} which is called HuntGPT, it was trained on an intrusion detection dataset and uses random forest to classify if logs are anomalies, the model used was GPT-3.5-turbo. The authors also use two frameworks, which give the model the ability to explain its results, this means that people who are not experts will be able to understand outputs. The model was tested on three recognized cyber security exams. Scores were ``between 72 and 82.5'' and their responses were informative and easy to understand. However, the model was not tested with event logs consisting of normal event logs and ones containing anomalies, which means that, although the model performs well at the exams and can explain suitably, we do not have any data that can highlight its performance anomaly detection. As a result of this, a future suggestion is to use the concept of explaining logs to people who are not experts in the field, in combination with accurately performing the task of anomaly detection. In addition,~\cite{hadadi2024anomaly} explore anomaly detection on unstable logs, to address limitations present in previous methods. This refers to logs that have changed over time in a certain application. They fine-tuned GPT-3 on a dataset containing both stable and unstable logs, as during their research GPT-4 was not yet available for fine-tuning. The results showed that their fine-tuned model had a higher F1 score and also found that it performed better at classifying unstable logs than the other model and it performed well when there were 30\% or lower unstable logs in the received prompt. The second test showed that, although GPT-4 is a newer model, the GPT-3 fine-tuned model outperformed both GPT-4 in the case of few-shot and zero-shot learning. Future work based on this article uses the method proposed with open-source models, such as LLama, as the authors mention that when fine-tuning GPT models they are ``disallowing any modification of their underlying architecture'' due to being closed source.

\subsubsection{Comparison}
As the two previous sections touched on solutions and work based on BERT and GPT, we will now look at articles comparing them to each other.~\cite{balasubramanian2023transformer} evaluate four GPT-3 models (Ada, Babbage, Curie and Davinci) on the task of anomaly detection. They use an open source event log dataset, to fine-tune the models, as well as provide a chat interface, giving users the ability to seek assistance from the model. The results showed that all models except Davinci, which received an F1 score of 0.4, achieved F1 scores above 0.9, and Ada was the superior model, achieving the highest score of 1. In addition, it is shown that all models except Davinci outperformed the models based on the BERT architecture. However, a limitation of this study is that it does not highlight that GPT-based models are online, which has the disadvantage of being a risk to privacy. Future work can look at developing models not based on the GPT architecture which can produce results that are equal or superior. Moving on,~\cite{mannam2023optimizing} evaluate the performance of three different classifiers using both GPT and BERT on four different datasets. The classifiers were the following machine learning algorithms: random forest, LightGBM and CatBoost. The results showed that the classifiers did not differ greatly in performance in all datasets, but all classifiers performed better and took less time to complete the task when using GPT. So, in terms of performance and accuracy, GPT is the superior model. However, a limitation of this result is that using GPT for event log analysis is a potential security risk, as event logs are sensitive data and the OpenAI API cannot be used offline; also, it sends data to their servers, whereas BERT does not have these two requirements and can be used offline. 

\subsubsection{Machine Learning Approaches}
We will now look at where researchers have used various machine learning processes with LLM-based anomaly detection methods.~\cite{10589612} present LogLead which covers three aspects of event log analysis: loading, enhancing, and anomaly detection. Loading refers to the retrieval of event log data. For this study, eight well-known datasets, including HDFS and BGL, were used. Enhancing refers to the process where parsing takes place, different parsing techniques have been used, such as drain, to break the logs down into a format which is easier to understand. The anomaly detection occurs by using multiple machine learning algorithms. This method is different from other research covered within this literature review, as others focus on one technique, for example, parsing or anomaly detection, but not both. LogLead was found to load the datasets faster compared to the other two datasets. One of the tools had a difference that was in single digits; however, the other had a difference which at times was more than a hundred million seconds slower compared to LogLead. A limitation of this study is that it requires different methods to analyse all event logs and over time will require updates incorporating new methods, as event logs will continue to vary in both structure and content. Future work can be developing a tool that will continue to be relevant without the need for an update when new structures and content of event logs emerge. 

Similarly,~\cite{li2024ids} present IDS-Agent, which is an intrusion detection system. Firstly, IDS-Agent extracts log data and prepossesses it, then uses various machine learning models to reach a result which is then summarised and outputted to users, using the LLM. The model can be configured as: GPT-3.5, GPT-4o-mini, and GPT-4o. They found that the best performance was with their GPT-4o model, then with their GPT-4o-min, followed by GPT-3.5, all models outperformed the base GPT-4o model. In addition, it was found that IDS-Agent detected zero-day attacks better than the other approaches as it had an average recall of 0.61, whereas the other approaches had an average recall of 0.41 and 0.47 respectively. A limitation of this study is that it utilises multiple machine learning algorithms for detection and does not highlight which is the best or if some are better at detection in certain scenarios. Thus, a suggestion for future work, based on this article is to look at which machine learning algorithms are the best to be used to classify logs as anomalies. 

A different approach is taken by~\cite{fariha2024log} who introduce a tool which utilises LLM APIs and machine learning for anomaly detection. The authors highlight the computation costs of fine-tuning and suggest that their method addresses these issues. There are three steps for their method. The first step is parsing where the GPT-3.5 API is used to take an event log input and convert it into regex form. The second step is processing the data using the Text-embedding-ada-002 API, specifically, the authors make sure that it is not too small or an amount which goes over the context window of the LLM. The third step does not utilize LLMs but instead uses a deep learning model, which was trained to construct normal event logs, it takes the processed data encodes it, then decodes it and attempts to reconstruct the log and if there are many errors in the process then the log is classed as an anomaly. The methodology of testing the model is to use the HDFS dataset and they conducted two tests, one was by using 80\% of the data for training and 20\% for testing and for the second 99\% for training and 1 was used for testing. On the first test, their method had an F1 score of 0.98 and on the second 0.998. Afterwards, they compared their results with similar methods which used LLMs for anomaly detection and their method had a higher F1 in comparison to them. Limitations of this study were that it requires human supervision and inputs, to successfully execute all the steps and that testing only took place with one dataset which does not give a general view of the proposed method’s performance. Regarding future work, research can be conducted to centralize the method into one LLM API, as it uses two APIs as well as a deep learning model. In addition, researchers can look at automating the tasks within the method, to reduce human intervention.

\subsubsection{In-Context Learning}
Proceeding, we now delve into the theme of using prompt-based methods,~\cite{egersdoerfer2023early} introduce a solution to address anomaly detection and issues regarding the low context windows of LLMs. Their solution does not require fine-tuning or RAG. It uses context creation which are prompts that allow the LLM to have a summary of its previous responses, addressing the context issue, and active analysis where it uses the summaries as knowledge to perform the task of anomaly detection. As the other solutions require training and fine-tuning, which has a large resource consumption, this method is a viable alternative. With regards to future work, more tests can be conducted and it can be compared to a larger amount of other solutions, to verify the results. Providing more data to assess the performance of in-context learning,~\cite{jin2024large} compare fine-tuning against in-context learning to discover which is superior in anomaly detection. Fine-tuning took place with freezing parameters and only allowing relevant parameters to be updated with new information. In-context learning took place using few-shot learning and zero-shot learning. The results showed that fine-tuning outperformed in-context learning at correctly classifying logs as normal or anomalies. Future work can look at improving the performance of LLMs at anomaly detection using in-context learning to make it a viable alternative to fine-tuning models. 

\subsubsection{Differing Approaches}
We now look at other research which provides solutions that do not fit into the categories of previous sections.~\cite{zheng2023logdapt} introduce LogDAPT which consists of two schemes, the first is Masked Language Modelling (MLM) which is ``to prevent replacing tokens with natural language tokens that are not present in log texts'' which, when masking tokens, replaces tokens with ones that are found in the data rather than with random data. The second scheme is called Span, and this is where tokens are masked together, ensuring that related parts are not masked separately, which could result in the model incorrectly understanding data that it was trained on. The results showed three trends: BERT was lower than the rest of the models and significantly lower in precision, both schemes generally performed around the same as NeuralLog and at times slightly better, and the span scheme performed slightly better and sometimes the same but never worse than the MLM scheme.  A limitation of this study was that it did not compare with a wide range of models to highlight how well it performs anomaly detection. As a result of this, future work can look at using the solution proposed in this paper and comparing it to other models.

~\cite{10414427} introduce a new method of detecting anomalies in event logs, called LogFiT. Previous methods such as DeepLog and LogBERT have been shown to stick to the patterns they were fine-tuned upon and thus when they encounter a different type or structure, they are unable to accurately classify them. In addition, these models have a limitation on the amount of input tokens that they can receive. LogFiT attempts to address these issues, as it does not require preprocessing of data before fine-tuning, researchers have recently found this is a reason for the issues, mentioned about previous methods. The reason for this is that the data on which the LLM is trained will be more specific after preprocessing, whereas unmodified event logs will contain more information. When new types or structures of event logs emerge, the LLM will have a higher chance of being able to understand them. In addition to this, due to being exposed to complete logs in training, it can identify anomalies more easily because it will see that the log received deviated from the patterns found in its training data. Addressing the issue of a limitation on the number of tokens the model can receive, the authors propose using `RoBERTa for sequences up to 512 tokens, and Longformer for longer log sequences'. The methodology of this paper is to train LogFiT, as well as DeepLog and LogBERT, on three publicly available event log datasets, then evaluate performance on anomaly detection. The results show a clear trend, LogFiT surpasses its predecessors by a large margin in F1 scores when trained with each of the three datasets. Regarding future work, it is mentioned that this study is restricted to anomaly detection and does not explore other aspects of event log analysis, for example explaining why an error has occurred, because of this, looking at using this tool in more broader event log analysis is a suggestion for future work. 

A different aspect is investigated by~\cite{guo2024logformer} with LogFormer, which improves on other models by retraining knowledge on different types of event logs. The authors highlight that in previous literature the methods struggled to comprehend different types of event logs. Their method has two stages, a pre-tuning stage, where the model is trained on common patterns anomalies follow, and adapter-based tuning, where the model is only updated with new layers, and previous layers are frozen to ensure that they are retained. LogFormer was found to be quicker to train as it took 29 minutes whereas other models ranged between 42 minutes to an hour and 48 minutes. With regard to accurate anomaly detection on a dataset (GAIA was used), again LogFormer performed the best with an F1 score of 0.93 and the other methods ranged between 0.30 and 0.88. The authors highlight that when words like ``error'' and ``failed'' are in the log data, LLMs will focus on them rather than understanding the logs. As a result of this, a suggestion for future work is to develop models that can recognise anomalies without relying on these keywords and instead utilise their knowledge. A specific type of logs is looked at by ~\cite{liao2024detecting} called WebNorm which detects anomalies specifically for web-based applications and if they are found, explains them. This method learns the processes used to produce web applications in their entirety, under normal circumstances, and uses this to define normal logs, if behaviour deviates from this pattern, then they are classed as anomalies. It was discovered that WebNorm is superior in its precision, recall, and F1 score compared to other methods. The F1 score (0.918) was 0.36 higher than the closest model to it, the other models ranged between 0.558 and 0.094 in their scores. However, the model also took a significantly longer time to train, as it took 1050 seconds and others ranged between 407 and 32 seconds. Future work based upon this article can look into implementing the method successfully but with reducing the training time compared to others and make a model which can produce the same results, not just specifically for weblogs, but others such as system and application logs.

Finally, we will look at cutting-edge research which attempts to address the limitations of LLM-based solutions.~\cite{10707565} attempt to address the high false positive rate of event logs. Previous methods falsely classify normal event logs as anomalies, when new logs or patterns that are not anomalies differ from their training data. In addition, two other issues with previous methods are mentioned, namely `log parsing errors'; this is where log parsers miss vital information, leading to the model not understanding the logs properly and `sequence modelling difficulty' which is where when there are changes in the sequence of activities, the models require retraining. The authors propose a method called LogRAG which uses semi-supervised learning, allowing the model to learn directly from the logs rather than from labelled data. In addition, it uses log tokens to split the log up into digestible pieces and log templates, which establish the structure of logs, to resolve log parsing errors. LogRAG detects anomalies by first checking if they match normal patterns in the training data, using a scoring system. The closer the score to the centre, the more normal it is, and the further away it is, indicates that it could be an anomaly. After this, a second check is made using RAG against anomaly event logs and if the input log has `a cosine similarity score exceeding 0.8' to the logs retrieved, then it will be considered an anomaly. The methodology of testing this method was to compare the anomaly detection result in the BGL data set with and without RAG. The results show that there is a clear increase when RAG was used compared to when it was not used. However, a limitation of this study is that if an event log is an anomaly, but RAG cannot find any matching logs, then it will still be falsely classed as normal behaviour.  Future work can address this limitation, as well as apply the use of RAG to other subdomains of event log analysis, although this method focusses only on anomaly detection. 

\subsubsection{Reducing Resource Consumption}
Moving on, addressing the limitation of large resource consumption~\cite{song2024audit} proposes an Intrusion Threat Detection (ITD) using LLMs called Audit-LLM, the novel aspect of this model is that it uses multiple agents. The first is the decomposer agent which sets out various steps that need to be completed as part of the ITD process, the second is the tool builder agent which builds tools that are required for all of the steps to be completed successfully, and the third agent is the executor agent, which uses the tools that have been developed to complete the tasks that are required. The results show that Audit-LLM significantly outperforms the other LLMs, for all three datasets. However, this method has the limitation of requiring an abundance of resources; if run locally, even in the case of APIs, the cost and time would still be a large amount. Regarding future work based upon this paper, the authors suggest to `reduce token usage', this will lessen the resource requirements, as the lower the number of tokens, the quicker it would be to train the model. In addition, they propose increasing the database of the model, with a larger variety of threats and the use of RAG to provide security professionals with relevant information on how to address threats highlighted by the system. A different approach to address resource consumption is taken by~\cite{yang2024anomaly} who fine-tune Llama2-7B and use it to train a smaller model, to reduce the size and the amount of resources required. The methodology was to test the base LLM, teacher and student LLM, at anomaly detection on a large dataset containing multiple types of attack. The F1 score for the base model is 0.72, the student was 0.87 and the teacher model was 1. As a result of this, future work could look at comparing the student model with other solutions and a larger variety of datasets. This will present an accurate picture of how well the model performs anomaly detection against other solutions and if it is the same or better, this can justify the slight drop in the F1 score of the original fine-tuned model. 

\subsection{Fault Monitoring}
To begin with, we look at~\cite{mudgal2024crasheventllm} who present the CrashEventLLM framework which aims to use LLMs, based on previous crash log data, to predict the time and types of future crashes. They fine-tune and use in-context learning to show the model how to analyse crash logs to predict future occurrences. They used a data set gathered from Intel where data was sent from the participating machines every 24 hours. They used two models Llama2-7b and Llama2-13b. The results showed that for time prediction both models received F1 scores of 0.39, for cause prediction the 7b model achieved 0.46 and 13b 0.34 and for full prediction F1 scores were just below 0.2. The results are significantly lower compared to other event log analysis tasks completed using LLMs, as a result of this, future work can fine-tune other models to attempt to improve results. In addition, the dataset for this study were limited, if a dataset like LogHub2.0 was used, this may yield significant improvements. Addressing some of these suggestions,~\cite{huang2024demystifying} propose LoFI which attempts to extract important data from logs. This prevents the need for security professionals to read raw event logs that consist of large amounts of noise. The model has three different components, preprocessed logs and extracting logs which focus on those that addressed faults, fine-tuning the model on a dataset generated by a company, and then in-context learning. LoFI responds with Fault-Indicating Descriptions (FID), which are descriptions of what the error was in the log, and Fault-Indicating Parameters (FIP), which are the area affected by the fault. They compared their method to two non-LLM-based methods and ChatGPT with zero-shot and few-shot learning. The results showed that LoFI achieved F1 scores ranging between 62.8 and 87.4 which is significantly higher compared to the other methods (GPT with few-shot learning, had the highest F1 score out of the other methods and it was 49.6). A limitation of this study is that the data used for fine-tuning were restricted. Future work can utilise a dataset that is not restricted to one company, which could improve results and allow the LLM to pick up on a larger range of faults.

Building upon previous studies,~\cite{shan2024face} propose LogConfigLocalizer which is different to other methods as it allows general users to address errors which occur in their systems. Raw logs are preprocessed using the Drain algorithm and logs specifically containing words associated with faults are extracted, and few-shot learning is used to teach the GPT-4 model. There are two main steps of this model: anomaly identification, where the LLM looks through the data to find logs that could be errors, and the second step is anomaly inference, where logs found in the first step are verified to be logs indicating faults. They tested the model by comparing its accuracy with others; the results show that ``has a mean accuracy of up to 99.91'' which was higher than ConfDiagDetector, which was non-LLM based. Some limitations of this study are the use of GPT, which is a security risk as sensitive data (event logs) would be sent to their servers, the method not being compared to other LLM-based methods, and an evaluation of how easy the outputs are to understand for non-specialist users. Future work can address this by utilising open source LLMs, comparing their methods with other LLM-based ones, and conducting evaluations of outputs to determine if they are suitable for users who are not security professionals. 

\subsection{Log Parsing}
\subsubsection{Early Work}
Parsing is a term used to describe the structuring of raw and unstructured event logs into a structured format. To begin, we will look at the earliest work that was found for parsing with LLMs. ~\cite{le2023log} evaluate the use of GPT for log parsing. The methodology of the study is to compare GPT-3.5.turbo with few-shot learning, against previous parsers, and to compare how different factors, such as the number of examples for few-shot learning or the complexity of prompts, affect performance. The results showed that GPT outperformed previous methods in message level accuracy (MLA) and edit distance (ED) and was second best in grouping accuracy, slightly behind Drain. Subsequent tests showed that the more examples used, for few-shot learning, the better the results and the more detailed the prompts are, the better GPT understands them and thus the results improve. A limitation is that the use of GPT has privacy risks, as data is sent and stored on OpenAI's servers, because of this future work can look at using LLMs which do not have this limitation. Similarly, ~\cite{ma2024llmparser} propose an LLMParser which uses few-shot learning. They tested their method on three LLMs T5Small, T5Base, LLaMA, and ChatGLM. An advantage of this method over others is that all LLMs used are offline, which addresses the limitation of online models that have security risks. The methodology of this paper is to compare the performance of the different models against each other and three other methods. The results show that the best performing model was LLaMa with the highest grouping (0.89) and parsing accuracy (0.96), it was also higher than the other methods to which LLMParser was compared, except for LogPPT, which had a higher grouping accuracy (0.92). However, a limitation of this model is that it requires updates over time, as the format of event logs changes, for example, with software updates, the structure and content of the logs will be different to what was used for LLMParser's few-shot learning. As a result of this, future work can address this limitation and take into account the changing nature of event logs without requiring updates, or in the case of methods using fine-tuning, retraining. 

\subsubsection{The use of different LLMs}
Selecting a specific model,~\cite{xu2024divlog} presents DivLog which uses few-shot learning with GPT-3 to perform parsing. The authors base their idea on previous research, showing that LLMs can learn from receiving examples of how to perform tasks. In the context of DivLog, they use the Determinantal Point Process as an algorithm to label event log data and use it to teach the model how to parse. To test DivLog, the authors compare it against other parsing methods including, the LLM-based, LogPPT. The results showed that DivLog performed better than the others followed by LogPPT. A limitation of this study is GPT-3 is online-based, meaning all the prompts are sent to and stored on OpenAI's servers, this causes security concerns. Thus, although this method produces results which are improvements, compared to prior methods, future work can look into using few-shot models with offline-based LLMs to build upon this study. Following a similar concept,~\cite{zhou2024leveraging} introduce LogBert which uses GPT-3.5 with few-shot learning to perform parsing to understand the ``semantic information of logs'' which the opposite is mentioned as a limitation of previous methods. In addition, it uses BERT to perform anomaly detection. The methodology of this paper is to compare LogBERT's performance on the HDFS and BGL datasets, as well as differing examples, to see how it performs as a parser. The results show that it can perform well, but slightly below the majority of the methods that it was compared to, as well as the performance improving depending on the amount of examples given in few-shot learning. Future work can build upon this study to produce LLM-based parsers which can produce results that are equal to or better than previous methods, potentially using other techniques such as fine-tuning or RAG, instead of few-shot learning.

Focussing on the performance of specific LLMs we look at a study conducted by~\cite{astekin2024comparative} who compared five LLMs, consisting of open- and closed-source models, to highlight which are better at log parsing. They evaluated with LogHub datasets and the use of zero-shot and few-shot learning. They found that, the open-source model, CodeLlama performed the best, the rest for few and zero-shot were GPT-3.5, Claude 2.1, and Llama 2. CodeUp was better than Zephyr, for few-shot and Zephyr was better with zero-shot. These surprising results show that CodeLlama, although open source and free, can outperform recognised and propriety models such as GPT-3.5 and Claude 2.1. As a result of this, future work can look at using CodeLlama for research, instead of propriety models like GPT. In addition, CodeLlama also comes without the limitation of privacy risks, as it does not require you to send data to their servers. In addition, research can investigate if CodeLlama is better suited for other event log analysis tasks such as anomaly detection or root cause analysis. However, one limitation with using open-source models, like CodeLlama, is that they either run locally or on a cloud server, both of which could have large costs.

\subsubsection{In-Context Learning}
Delving into zero-shot learning,~\cite{zhi2024llm} propose YALP as an alternative to fine-tuning GPT models for parsing due to the large amount of resources required and costs. Their method uses zero-shot learning on GPT-3.5, where logs are preprocessed and matched against templates; if there are no matching templates, then a new one is produced. The model is evaluated on the LogHub-2.0 datasets against four other non-LLM-based parsers and the base GPT-3.5 model. The results show that YALP achieved an average grouping accuracy of 0.92, parsing accuracy of 0.72 and edit distance of 3.53. The grouping and parsing accuracy results were the highest, the edit distance was the lowest compared to the other methods and the base model. However, a limitation is that it requires templates that are generated following previous prompts and the model is not compared to others that use fine-tuning. Future work based on this paper can compare zero-shot learning to fine-tuned models, to build upon this study. Providing an answer to this suggestion,~\cite{yu2024loggenius} propose LogGenius which does not require fine-tuning or few-shot learning and instead uses zero-shot learning. Instead of the LLM directly being trained or learning, they use non-LLM-based, unsupervised log parsing methods (Brain, Drain, Logram, Spell, AEL and Lenma). They conducted three tests: comparing using LogGenius with each method and against the base methods, the same test but with unseen logs and against GPT-auto (zero-shot learning using prompt engineering to produce results), and these were all measured on the percentage of parsing accuracy. The first test shows that Brain, Drain and AEL improved when used with LogGenius, whereas the others generally stayed the same. In the second test AEL, Lenma generally stayed the same, and the others had slight improvements. In the third test, all methods with LogGenius, except Lenma, outperformed GPT-auto. In addition, Brain produced the best results of all the methods that LogGenius used. Future work can look at using the LogGenuis method, with open-source LLMs, as in this study GPT-3.5-turbo was used, which has the limitation of security issues.   

Moving on, to look at in-context learning in a more general sense.~\cite{jiang2024lilaclogparsingusing} introduce an event log parser using LLMs, known as LILAC. The authors highlight, before their study, that the available models had various limitations, some of which include large amounts of computing resources required for fine-tuning, inconsistent outputs and the large amount of event log data produced by systems and LLMs having restrictions on the amount of data that they can process. The proposed method has two main components. The first is an ICL-enhanced parser, which instead of fine-tuning, uses in-context learning to teach the LLM the process to correctly parse the log. The second is an adaptive parsing cache, which is where the received event log is matched, using k nearest neighbours, to match the closest to it in terms of similarity. The advantage of the solution is that if an event log has already been parsed, there will be a template stored, and both will be matched, saving the need for re-parsing. The selected model was GPT-3.5-turbo and they `set its temperature to 0', which allows it to produce the same responses. The methodology for testing was to compare their model with its predecessors. The first test showed that it was superior by a large margin in parsing, except from Drain which was at times only slightly less accurate, but the authors mention that this is `due to the imbalanced frequencies of templates in log datasets'. The second test found that the use of a parsing cache improved the model performance by almost 50\%. The third test found that LILAC was able to integrate with models outside of ChatGPT with only a slight decrease in accuracy. The final test showed that the model was able to process a large amount of event log data, faster than all author LLMs against which it was compared, except for Drain, which was said to be `the most efficient syntax-based log parser currently available'. Future work based on this article can address the issue of successfully and efficiently inputting event log data into a LLM, as they are generated, the authors highlight that this is a limitation which is yet to be addressed. 

In addition,~\cite{wu2024log} introduce AdaParser, which attempts to address the limitation of log parsers including LILAC at logs which change over time. As software and systems are updated, the log formats also change, which means that previous methods could not be used after a certain amount of time. As updates take place frequently this is a major limitation because after a short amount of time parsers would either need retraining or become redundant. AdaParser checks for templates if none are generated. Self-correction makes it unique and it is where, if the template does not match the pattern of a log message it is ``fixed'' which allows the LLM to correctly understand it. The examples given in the paper show that if an LLM interprets related parts of a log message separately, the self-correction method will show the LLM that they are not matching until it understands that they are related. The methods of testing AdaParser were to compare its performance with those of LILAC and other log parsers, which were not based on LLMs, on Loghub-2.0 datasets and in zero-shot scenarios. The results showed that AdaParser achieved results on Loghub-2.0 datasets, which were higher than both LILAC and Drain which were the log parser that performed the best before it. In addition, it was also superior in zero-shot scenarios and improved performance with a variety of LLMs. However, although it improved performance in zero-shot scenarios, the results were ``significantly inferior to that in few-shot scenarios'' which means that it would still be usable but would have a decrease in performance over time. As a result, future work could address this and produce a method which can maintain performance in both few-shot and zero-shot scenarios. 

Building upon previous studies, ~\cite{vaarandi2024using} propose LLM-TD (LLM Template Detection). The limitations that it attempts to address are the assumption that all logs that conform to a template have the exact amount of words, as well as privacy concerns with online models. They use in-context learning to address the high resource consumption and costs of fine-tuning, and their method allows them to analyse multiple logs and find multiple templates at a time. The method was tested using five Linux-based event log datasets and compared three LLMs (OpenChat, Mistral and Wizardlm2) and Drain. The results showed that OpenChat was the best performing LLM, which produced results that were close to drain performance, received higher F1 scores compared to it, and was the only LLM that was able to analyse all the data. Mistral, although it could not complete the analysis, achieved higher F1 scores than Drain. Wizardlm2 was the worst performing LLM and achieved scores that were less than Drain. Some limitations of this study are that it does not take into account the limited context window of LLMs, which would restrict the amount of logs that models can answer at a time and there is no comparison to methods using in-context learning on online models. Due to this, future work can investigate the method and the impact of the limited context window and compare models based online and offline on their performance with in-context learning. Adhering to the concept of addressing limitations,~\cite{zhong2024logparser} propose LogParser-LLM which addresses the limitations of a lack of labelled event log datasets and the large number of resources and costs associated with fine-tuning. They use in-context learning, since LLMs have a ``deep semantic understanding and the capacity to generalise across different log formats'', which they attempt to take advantage of, to avoid the limitations mentioned previously. They evaluated their method against other log parses and evaluate which is the best model with LogParser-LLM. The results show that on average LogParser-LLM had an average F1 score of 0.91 grouping accuracy and 0.81 parsing accuracy, which was higher than the other models against which it was compared. In addition, the model performed best with GPT-4. A limitation of this study is the privacy issues which come with using GPT-4, as it is online, because of this future work can evaluate using a method with is closed source and produces equal or superior results.

Proceeding, we look at solutions that do not require labelled data and are also not based upon in-context learning.~\cite{huang2024ulog} propose LUNAR, which is an unsupervised LLM parser, which does not require labelled data, unlike other methods. First, they separate logs into different groups, then rank which parts of the logs are most unique and relevant for analysis, and finally, a prompt is made describing how to perform the task, and then the parsed logs will be the LLM's output. The method was evaluated by comparing it to other methods on a large variety of datasets using metrics, a set of previous researchers, to evaluate performance in parsing. The results showed that LUNAR outperformed other methods in all metrics but in general was only a slight improvement over LILAC. Future work can investigate improving performance to increase superiority over LILAC. Another approach is taken by~\cite{ma2024librelog}, who attempts without requiring labelled data, called OpenLogParser. They address the issue of security by using Llama3-8B rather than a GPT-based model, avoiding the limitation of it being restricted to online use with privacy risks, and choose it as it is shown to have the best performance with this method compared to other LLMs. The method uses RAG, and then the retrieved logs are scored and logs which have scores showing a range of content are retrieved, then a prompt is used to specify how to perform the task, and finally, the parsed logs are stored, which saves the need for multiple queries to the LLM. The model was tested using Loghub-2.0 datasets and compared its performance against four other methods, two of which were LLM-based, LILAC, and LLMParser. OpenLogParser performed the best, followed by LILAC then the two other methods, and the lowest scores were found in LLMParser. A major advantage of this method, compared to the others, is the reduction in resources because of a reduction in queries to the LLM due to the memory function. Future work based on this article can assess the model performance on logs that differ over time, ensuring that new logs are retrieved, as there are limited amounts of event log datasets, due to the fact that they are sensitive data and when log formats change, there is a high chance that they may not appear in any dataset.

\subsubsection{Fine-tuning}
Shifting our attention to fine-tuning,~\cite{chen2024high} attempts to address the lower efficiency of previous deep learning parses with Hooglle. Their method first performs offline fine-tuning on previous logs to make sure that future outputs adhere to their format. Secondly, online parsing takes place where each log is analysed individually, if it matches previous templates, then they are assigned to it, if not,, then a new template is generated. To test Hooglle it is compared against four other log parsers on a variety of datasets. The results showed that on average Hooglle performed the best, throughout all datasets, its parsing and group accuracy only ranged between 0.9 and 1. Although the results are high, a limitation common with other LLM-based parsers is security issues as they are online-based and this can be addressed in future work.

\subsubsection{Solutions for previous limitations}
To conclude, we look at solutions address papers which address some of the limitations mentioned frequently in previous sections.~\cite{xu2024help} propose HELP which is an online parser which attempts to address the high costs found when using other LLM-based parsers, which in the paper is highlighted to have the potential to ``cost hundreds of thousands of dollars a day''. First, the logs are turned into vectors based on their similarities to prevent individual parsing, then few-shot learning is used with Claude-3.5-Sonnet teaching it how to perform parsing and templates are updated over time, to allow similar logs to continue to be grouped. The method was evaluated on Loghub-2.0 datasets against other parsers. The results showed that HELP achieved the second-best results, but in all metrics, it was slightly behind LILAC. Some limitations of this study are its reliance on GPT and Claude which have the risk of security concerns, compared to open-source models, which do not require data to be sent to and stored on servers. Future work based on this article can address these security concerns and the development of a model based on this concept that can surpass or equal the performance of LILAC. Moving on, ~\cite{mehrabieffectiveness} attempts to address the limitation of security risks with the use of closed source models, for example, the GPT series and Claude, in log parsing. For their evaluation, they fine-tune Mistral-7B, with various well-known event log datasets, and compare it, as well as the base model, with in-context learning, against GPT-4 also with in-context learning. The results showed that the fine-tuned Mistral-7b model was slightly better than GPT-4 with few-shot learning, and significantly better compared to zero-shot learning. Mistral-7b with few-shot and zero-shot learning produced the lowest results. However, a limitation of this study is that the fine-tuned Mistral-7b model is only compared to GPT with in-context learning and research suggests that GPT parses better when fine-tuned; this means that we cannot conclude that Mistral-7b is the superior model at parsing. As a result of this, future work can compare Mistral-7b and GPT with fine-tuning to verify and build on this study. 

Another issue is investigated by~\cite{astekin2024exploratory} who evaluate how ``non-determinism'' affects log parsing with LLMs as previous research showed this was present with code generation. The authors highlight their concern that if this is the case with parsing as well, the results from previous LLM parsing studies may not be accurate. Non-determinism, in the context of LLMs, is where the response differs for the same prompt. The methodology was to use zero and few-shot learning to teach the model how to respond, and then a prompt with log data is given fifty times. This study is conducted on two closed-source LLMs (GPT-3.5 and Claude 2) and four open-source (Llama 2, CodeLlama, Zephyr Beta and CodeUp). The results showed that GPT-3.5 had the least amount of non-determinism, followed closely by CodeUp, then Claude 2, CodeLlama, Llama 2, and Zephyr had a lot of non-determinism. In addition, when the temperature was closer to 0, as expected, the prompt was more deterministic, but interestingly this was not always the case. Future work based on this article can work on producing models where the responses are deterministic (stay the same). Proceeding to address another limitation,~\cite{xiao2024stronger} which aims to address resource consumption (due to training) and large costs (due to an abundance of calls to the LLM to complete tasks), with their method LogBatcher. They first group log data, then logs that differ from each other are used as a prompt, compared to other models which use labelled data used for few-shot learning. This is an advantage because if the logs are similar, then the LLM will only be queried once. The method was tested on multiple event log datasets, against other methods, including LILAC. The results show that LogBatcher performs better than all other methods, including LILAC. However, a limitation of the method is that in the future, log formats will differ in their structure and content due to software and system updates, resulting in the method potentially not understanding them. Future work can look at addressing the issue of LLM-based parsers requiring updates or becoming redundant as time progresses and log formats change.

\subsection{Root Cause Analysis}
To begin, we look at the first paper on the subject of root cause analysis using LLMs.~\cite{ahmed2023recommending} study the use of fine-tuning for root-cause analysis. The main aim of the paper is to test fine-tuned GPT-3 models (``Curie, Codex, Davinici and Code-davinci-002'') by performing root cause analysis and then generating relevant suggestions based on this information. The dataset used contains data from a large variety of incidents which took place over six and a half months at Microsoft. The researchers conduct various tests including evaluating whether the root cause was identified, recommendations, comparing to zero-shot, and human evaluation of outputs. For all of the tests Curie, Codex and Davinici performed relatively the same and Code-davinci-002 performed slightly better than the rest. In addition, all models performed better with fine-tuning than the base models (with zero-shot). However, although security professionals generally approved of the output, they mention that at times they could be incorrect and require manual verification to mitigate this. Also, the use of GPT poses security risks as it is a closed-source LLM and requires sending data to their servers. Future work can look into improving the accuracy to make it perform automatically without the need for verification and looking into open-source LLMs. Addressing this limitation,~\cite{mandakath2024root} analyse the analysis of the root cause fine-tuning Mixtral-87B and MistralLite. Their methodology was to use the base GPT-4 model to extract the root causes, and the researchers used this to compare the performance of the two fine-tuned models. The results surprisingly showed that MistralLite which is smaller in size performed better, in addition, ChatGPT was slightly worse in performance compared to it. This could be because fine-tuning a larger model is more difficult than a smaller one due to more parameters that need to be changed. However, although GPT performed well, security experts had around forty-one percent confidence in the outputs it produced. Future work based on this article can look at evaluating root causes with security professionals to see if the LLM's responses meet their standards.

Proceeding,~\cite{10.1145/3663529.3663841}, who present a Root Cause Analysis (RCA) system, called ReAct, using LLMs. These systems are important, as they explain why faults occur and the information gathered can be used to prevent future occurrences. RCA systems, utilize a wide range of information, including event logs and documents, which describe previous work. The authors touch on fine-tuning and its significant costs and high resource requirements and propose to RAG to retrieve data, then utilize few-shot learning, to teach the system the best ways to respond. The methodology of this study is to use the GPT-4 model, as it was the most powerful model when the authors complied with this paper and it can process larger amounts of tokens compared with other models. It was evaluated by performing and assisting RCA tasks when implemented into an Azure department within Microsoft. The results showed that the model was able to complete generic tasks, however, on more complex tasks it made severe errors. In addition, the experiments were only held within Microsoft’s environment and other companies will differ in their requirements and how they operate and how their systems work meaning that the results gathered within this paper may not accurately represent how the system works in a general sense. Although reducing costs and the number of resources required is an important consideration, all security-related models need to ensure accuracy and if there are many hallucinations and false positives, organizations that use such models will be less secure. Future work based on this article can look at improving the performance of the model on more complex tasks as well, as a wider range of environments which will address the limitations and provide a clear image of how accurate the model is. Building upon this study,~\cite{wang2024rcagent} address previous methods not using features of LLMs which allow it to understand tasks and instead use manual processes, as well as being based on open source LLMs posing security risks. Their method, RCAgent, first, uses OBSK which allows only important information to be analyzed reducing the number of tokens which address the limited context length of LLMs. Secondly, tools are used which capitalize on LLMs being able to reason and understand tasks, specifically information gathering tools are used to only gather relevant information and analytical tools allow it to analyze the logs. Thirdly, stabilization takes place which ensures that the format of outputs is structured and organized (stable). Finally, the LLM evaluates itself to ensure that its response is accurate. The model is evaluated by comparing it to ReAct and it is shown that RCAgent is a significant improvement, as it obtains a pass rate of 99.38 compared to ReAct which had 86.33. A limitation of this method is the amount of resources that would be required, as it is run locally, powerful hardware is required. To address this, future work could look at using closed-source APIs, which share the advantage of being secure, but do not require as many resources as RCAgent.

Moving on, we look into studies that specifically evaluate the use of few-shot learning as an alternative to methods, such as fine-tuning, mentioned previously.~\cite{chen2024automatic} propose RCACoPilot which aims to minimise the need for manual root cause analysis by allowing the LLM to complete the task automatically. This approach uses a few-shot learning with GPT-3.5 and 4. Their approach consists of two main objectives, in the first, inputted logs are parsed and then matched against a type of alert. The second part is where the LLM uses the information obtained from the previous part to predict the root cause of an incident. They compare their method using both versions of GPT with three fine-tuned LLMs. The results show that RCACoPilot achieved the highest F1 scores (0.761 and 0.766). However, the F1 score was not significantly higher; in addition, the average time taken was also the same with both GPT iterations. This means that GPT-3.5 could be seen as the better option as the difference is insignificant, but it costs less to use. A limitation of this study is the use of GPT which is a concern as the data will be sent to OpenAI's server. As a result of this, future work can look at using alternatives that do not have this limitation. Similarly,~\cite{zhang2024automated} addresses the large resource consumption of fine-tuning and sees if in-context learning is a viable alternative in the case of root cause analysis. Their method gathers log data which has been obtained from a cloud company over nine months, then in-context learning takes place to summarise the root causes of all the incidents, embedding vectors are then generated and finally, Faiss library is used as a retrieval index to retrieve the data (embeddings) when relevant and needed. The model is then compared against a fine-tuned GPT-3 model and four base LLM models. The results showed that GPT-4 with in-context learning performed the best ``by an average of 24.8 across all metrics''. However, a limitation is that at the time this paper was compiled, only GPT-3 was available for fine-tuning, as others like GPT-4 are now available, the results may be different because other models, with fine-tuning, could perform better than GPT-4 with in-context learning. As a result of this, future work can be compared with a larger range of fine-tuned LLMs to establish whether it is a superior method in the case of root cause analysis.

\subsection{SIEM}
To begin, we look at an early study from~\cite{sultana2023towards} that evaluated developments that could be addressed for cyber security-related processes using LLMs. The methodology was to look at the literature before it and use this to reach conclusions. As the study was early in the field, there was not much published, some included the early articles on anomaly detection discussed in this survey. The main suggestion based on their research, which also closely links to event log analysis, is to work on developing datasets, due to the small amount available in the cyber security domain. The reason why this is important for event log analysis, in particular, is because event logs are sensitive due to containing personal information and activities that a user would do on their devices. Future work can investigate the development of large event log datasets without compromising users' privacy. Continuing, ~\cite{kaheh2023cyber} present Cyber Sentinel which uses GPT-4 to aid cyber security professionals who manually analyse data, such as logs, to prevent attacks. The cyber sentinel addresses this by performing tasks and providing recommendations that take advantage of the LLM abilities. They use Elasticsearch to search the IoC database to retrieve relevant logs which will assist in performing tasks, and Wazuh for intrusion detection. They use prompt engineering to show the LLM how to respond and ensure that it is within the domain. A limitation of this study is that it lacks a comparison with other methods, for example, LLM-based intrusion detection systems like HuntGPT, which could highlight its advantages as it is not restricted to a certain aspect of event log analysis. Another early study,~\cite{yao20235g} aims to address two main limitations: the large amount of data needed for fine-tuning and security issues associated with using open source providers, specifically focused on security alerts for 5g. They used GPT-4 to produce the dataset for fine-tuning due to its high performance and then fine-tune Llama2 which is closed source to address the second. They evaluated their method on a virtual machine and showed that it has a correlation accuracy of ``up to 89.5'' which is shown to be close to GPT's performance, but without limitations mentioned previously. A limitation mentioned by the authors is that attackers have the potential to attack or influence the model results or find ways to bypass it. Future work can address this limitation by adding security to the model and measures to prevent attackers from influencing or bypassing the model.

Proceeding,~\cite{ahmed2024prompting} have a unique approach in which they evaluate GPT to provide CSC suggestions, providing organisations with the best practices to ensure that they are secure on a technical level. Their method uses few-shot learning and chain-of-thought reasoning to teach the model how to respond, and they use previous prompts as a knowledge base. An example given in the paper is for deleting or deactivating an account if it is not used for a certain amount of time. The LLM will receive event logs, look for logon and logoff event IDs, and then suggest that the account should be deleted or disabled if they have not been found in the prior 45 days. Their method was evaluated using both an LLM and human evaluation. The results show that their method achieved high correctness scores that, in general, were between 8 and 10 and this was the case with both human-based and LLM-based evaluations. In addition, they also find that, as expected, GPT-4 outperforms GPT-3.5 in performing the task. Some limitations that the authors mention of this study are the potential for hallucinations and that the prompts used may not necessarily work if this method were conducted in another LLM. Future work based on this article can investigate chain-of-thought reasoning and few-shot learning on other LLMs to further evaluate its performance.

Shifting the focus, we will now analyse articles that evaluate models over time. Beginning,~\cite{karlsen2024benchmarking} evaluate the use of five models ``BERT, RoBERTa, DistilRoBERTa, GPT-2, and GPT-Neo'' and compare their performance in the analysis of logarithmic data. They propose the LLM4Sec framework firstly the inputted logs are split for training (fine-tuning the models) and to test the performance, after this, the results are then put in a visual format which allows security professionals to assess performance. The results show that DistilRoBERTa performs the best with the base model, achieving an average F1 score of 0.908 after fine tuning 0.998. This was followed by the following models in the case of fine-tuning RoBERTa (0.997), BERT (0.994), GPT-Neo (0.987) and GPT-2 (0.953). The results show that fine-tuning significantly improved the scores for all models, which made them all achieve F1 scores above 0.9. However, DistilRoBERTa (which was the model with the smallest parameters) made an insignificant improvement after fine-tuning. In addition, the base model achieved scores that were better than all other models even after fine-tuning. Future work based on this study can investigate this trend, as it would usually be expected that the larger the model, the better it will perform at analysis, but this study suggests the opposite. Subsequently,~\cite{jonkhout2024evaluating} evaluated the performance of different LLMs to respond to alerts for cyber security incidents. Security professionals who, due to the high frequency of less critical/false positive alerts, may not pay enough attention due to ``fatigue'' resulting in potentially critical alerts being overlooked. They use prompt engineering to teach the models how to analyse security logs and determine if they are an actual attack addressing the issue mentioned previously, as this allows security professionals to respond to actual alerts rather than low-level or false positive alerts. Their methodology is to compare three LLMs: Llama3 (8 billion parameters), Qwen (4 billion parameters) and Phi3 (14 billion parameters), all three models are open source. The results show that Llama3 (2.06) performs slightly faster than Qwen (2.76) and both are significantly faster than Phi3 (19.53). In addition, all free models have virtually the same average accuracy varying between 0.553 and 0.556 for LLama3 and Qwen respectively to PHi3, which was 0.573. This suggests that although a model has larger parameters as in the case of this study, it may not result in any noticeable improvements. A limitation of this study is that all the models compared were open source. Future work can build upon this study by comparing a large number of LLMs, including open source, and looking at other methods such as fine-tuning to improve scores as the accuracy of all models in this task is low. 

Concluding~\cite{hermann2024gpt}, evaluate GPT on the analysis of logs and, when necessary, classify them worthy of the attention of security professionals. The authors highlight that security professionals analyse a large number of logs, most of which contain normal activity, and due to fatigue critical logs may be missed, which poses a security risk. The methodology of this paper is to compare the base 3.5 GPT model against a fine-tuned version, with both, using prompt engineering to specify the required format. The test was carried out using the ATLAS dataset. The results clearly showed that the fine-tuned model performed better as F1 scores ranged between 0.73 and 0.83, compared to the base model, which ranged between 0 and 0.001. A limitation of this study is the restriction to only one dataset of logs and the comparison using one version of the GPT model as well as not comparing to other LLMs. Future research can look into using a larger range of datasets and LLMs to compare with. Addressing the limitations mentioned in the previous two articles,~\cite{oniagbi2024evaluation} conduct a comparison of the following models ``GPT-4o and GPT-3.5, Meta Llama 3, Mixtral 8x22B, and OpenHermes 2.5 Mistral 7B'' at reviewing alerts (which contain the corresponding logs) to determine which requires attention. The methodology is to use prompt engineering to allow all of the models to understand how to perform the task. The results show that the best performing model (based on the F1 score) was Llama 3 (45.3) followed by GPT-4o (42.4), Mixtral 8x22B (41.5), GPT-3.5 (36.2) and OpenHermes 2.5 Mistral 7B (31.1). A limitation of this study is that fine-tuning was not used which could have improved results, as even the highest performing model, Llama 3, has low scores. Thus, future work can build on this study and implement fine-tuning.

\subsection{System Health Monitoring}
There are currently no papers found, during the compilation of this survey, where event log analysis using LLMs has been used specifically to monitor the performance of devices.

\subsection{Threat Detection}
\subsubsection{Use Cases}
To begin, we look at survey articles related to the topic,~\cite{chen2024survey} survey a general sense of all applications of threat detection using LLMs. Their methodology was to use literature to identify the progress up to the point the article was written and then make suggestions for areas that can be investigated in future research. As this was an early piece of work there were only a few papers written specifically to event logs which mainly used GPT-2 but there were some such as LogGPT, mentioned previously in this paper, which began to use GPT-3. There was a lack of papers focusing on specific issues such as comparing fine-tuning and in-context learning and how accurate were the models and methods to improve this such as using RAG. A survey is also conducted~\cite{khanllms} about malware, but it does not directly address event logs. However, papers are evaluated relating to topics where event log analysis is crucial, such as detecting ransomware. The authors highlight, that although current solutions perform well, they are vulnerable to data differing from that which was used for training, requiring retraining. In addition, malicious actors could change existing code which could result in the models not understanding the new changes as, even with small changes, the content within logs could change drastically. Based on this, future work can focus on developing models which do not rely on input data, to match training data, which would address new logs or those which have changed from their original format. Finally,~\cite{patil2024leveraging} evaluate the use of LLM as a tool for cyber security in cloud-based networks. Their methodology was to conduct a thorough review of literature and industry applications. They highlight two examples where LLMs were used for assisting tax and for a retail company, resulting in an improvement of efficiency in both cases. Another example is highlighted when three companies used Microsoft's LLM, CoPilot, for cyber security and the authors mention that ``the accuracy level started at around 50\% and rose to within the range of 75-95\%'' after the use of CoPilot. Despite this, it is mentioned that current implementations have the risk of hallucinations and security risks. The authors suggest, to address security risks, applying least privilege to LLMs, but this could reduce their impact as a security tool, as they will not have access to all data and as a result can not be a complete security tool. Future work based on this article can research and provide suggestions to reduce hallucinations, as well as produce an LLM-based tool that does not require least privilege with the development of secure models.

\subsubsection{Machine Learning}
Proceeding, we look at approaches which use machine learning,~\cite{ferrag2024revolutionizing} introduce SecurityBERT to detect various types of threats affecting IoT devices. They use the PPFLE algorithm which takes raw logs as well as other relevant forms of data and converts them into a format, which is similar to the English language which the LLM, in this case, BERT, could understand better After this it is tokenised and then used for training. The method is compared to machine learning and deep learning models and is shown to achieve 98.2\% in accuracy which was higher than all of the other models that it was compared against. A limitation of this study is that is specific to IoT devices. Future work can look at the implementation of this method for a range of devices and can investigate the use of the PPFLE algorithm with other applications of event log analysis as this study represents that, in the case of BERT, it managed to achieve highly accurate results. A similar approach is taken by ~\cite{hasan2024distributed}, who investigate the use of LLMs for threat detection for edge devices. Their methodology uses machine learning algorithms on edge devices, to detect threats and uses an LLM (GPT API) to act as the main edge server and in-context learning is used to teach the model how to detect and respond to prompts, future prompts are used to ensure knowledge is kept up to date. It is mentioned that the connection to the LLM is ``through a high-speed network'' which allows it to analyse the log data and respond quickly if threats are detected. This was an early study, the proposed solution was not implemented or tested and the authors mention that in future work they will implement the system. Some limitations of the method are the large costs associated with calling the API to analyse the information and also security concerns with using GPT due to data being stored and processed on OpenAI's servers.

\subsubsection{Comparison}
Moving on to papers which compare various models, we begin with~\cite{sanchez2024transfer}, which compare six LLMs (BERT, DistilBERT, GPT-2, BigBird, Longformer and Mistral Small) on threat detection after being fine-tuned on a syscalls dataset, which are calls to the operating system. They use the MalwSpecSys Dataset, which contains large amounts of malware-based data, to evaluate the performance of the different LLMs. The results show that DistilBERT performed the worst followed by Mistral Small, BERT and GPT-2, Longformer performed the best as it was slightly better than BigBird. The results in general show that the smaller the context size the worse the model performs, which is expected. However, an exception was that Mistral Small, performed the second worst, although it was the largest model and had a close to double context size to BigBird and Longformer. Future work based on this study can investigate further into why Mistral Small performed worse despite having a superior context size. Likewise,~\cite{houssel2024towards} evaluate the use of Llama3 and GPT-4 on network-specific threat detection. Their methodology was to evaluate their performance with zero-shot learning (the base model) and with two fine-tuning methods (KTO and ORPO) on datasets comprised of network event logs. The results for zero-shot show that GPT performs slightly better, as on the first dataset Llama3 has an average precision of 48.56, GPT-4 had 50.85 and on the second dataset, Llama3 had 49.66 and GPT-4 had 50.25. Fine-tuning only took place for Llama3 and produced results that were the same as GPT and slightly better than the base model. A limitation is that the two fine-tuning methods do not change all of the parameters or a large amount which could explain why there is a close to insignificant improvement after fine-tuning took place. Future work can be evaluated using fine-tuning methods which change a larger amount or all parameters of a model.

\subsubsection{Fine-tuning}
Continuing on the theme of fine-tuning,~\cite{novaklightweight} evaluates the use of fine-tuning ChatGPT for ransomware analysis to prevent a reduction in performance for example reducing network speed. The logs gathered, used as the dataset for fine-tuning, were from industry-based computers to represent normal behaviour and permissions and the model would detect ransomware as when it occurs the logs will differ from this normal behaviour. To ensure the model was up to date, it was fine-tuned as ransomware and systems evolved. The model was tested to show its performance in different scenarios which were with different types of ransomware and differing network loads. The results show that the model had an accuracy above 90\% and false positives only ranged between 1\% and 2.5\%, regardless of the type of ransomware and load of the network. In addition, when the model was used most results show that it had little to no impact on systems and the network, highlighting that it achieves the objective the researchers intended. A limitation of this method is that it requires fine-tuning whenever there is an update, which will take a large amount of time and resources, if it does not take place then the security could be compromised. To address this, future work could look at using automated methods such as RAG which will allow the model to be updated automatically without the need for manual fine-tuning. 

Addressing some of these suggestions,~\cite{lanka2024intelligent} propose using LLMs to analyse honeypot logs to prevent threats. Their method gathers data and uses k means clustering in combination with RAG to identify if logs are normal or malicious. The method was evaluated over 7 days in an organisation and concluded that the method was able to correctly identify attacks at a fast speed comparable to cloud-based solutions. However, limitations are the costs associated with using GPT's API, the hardware required to run the model and security issues with using GPT. Future work can build upon this study by investigating alternative LLMs which are secure and methods to decrease the resource requirements. A similar approach is taken by~\cite{rahmani2024integrating}, who propose an LLM-based threat detection solution. They fine-tune Llama-3-8B-Instruct with the CIC-IDS2017 dataset and implement various RAG procedures. The model was evaluated on the same dataset and achieved an accuracy of 96.85\%, this was compared to other methods such as SecurityBERT which achieved a slightly higher accuracy of 98.2\%. A limitation of this study is that the dataset on which the model was evaluated was the same one used for fine-tuning, this means, as other research into LLMs has discussed, patterns or the data itself could have been memorised by the model in fine-tuning. This could result in scores which may not be an accurate representation of the model's accuracy. Future work based on this article can look into comparing models to assess which would be the best for threat detection.

\begin{table*}[hbt!]
\centering
\begin{tabular}{|l|p{11cm}|}
\hline
\textbf{Category} & \textbf{Summary Findings} \\
\hline
Anomaly Detection & Several methods including BERT-based solutions, GPT series, deducing which models are the better for the task, improving accuracy and efficiency of tokenisation, training and reducing resource consumption, machine learning, in-context learning and RAG. \\ 
\hline
Fault Monitoring & Crash prediction using LLMs and extraction of fault-indicating information from logs for failure diagnosis. \\
\hline
Log Parsing & Early work on log parsing using LLMs, use of different LLMs for parsing, in-context learning for parsing, RAG for parsing, fine-tuning for parsing. \\
\hline
Root Cause Analysis & Early work on root cause analysis using LLMs, few-shot learning for root cause analysis, fine-tuning for root cause analysis, RAG and addressing high resource consumption issues. \\
\hline
SIEM & Early work on SIEM using LLMs and evaluating models over time, reducing resource consumption and explaining analysed logs to non-experts. \\
\hline
System Health Monitoring & No specific research was found on system health monitoring using LLMs. \\
\hline
Threat Detection & Threat detection using LLMs, machine learning approaches for threat detection, RAG for threat detection, fine-tuning for threat detection, techniques to address concept drift. \\
\hline
\end{tabular}
\caption{Summary of findings from the survey on event log analysis using LLMs.}
\label{tab:summ}
\end{table*}

\subsubsection{Concept Drift}
Concluding,~\cite{sundin2024ai} aims to address ``concept drift'' which is a term used to describe the changing structure of logs over time and has been mentioned, in previous work, as a major limitation for event log analysis using LLMs. In tasks like anomaly detection, if the structure of a log changes, models could interpret what is a normal log as an anomaly because it differs from the patterns that it was trained on. To address the issue of context drift, they produce embeddings for words that are similar to those contained within logs. An example of this, for the word execute, could be run which would be generated, if this word is changed in future logs, the LLM will not class it as an anomaly. The approach was tested on the BGL and HDFS datasets, with and without the proposed solution to concept drift. The results showed that after concept drift their approach suffered a minimal reduction in accuracy as it achieved F1 scores of 0.831 (HDFS) and 0.871 (BGL) compared to without their method which had F1 scores of 0.931 (HDFS) and 0.952 (BGL). In addition, the scores were slightly higher than the previous approach. A limitation is that the data obtained for this study were from one company and could differ from or do not contain data that would be present to others who could be based in different sectors. Future work based on this article can investigate tokenizing ``parts of words as separate tokens'' as the authors suggest that it could improve the results.

\subsection{Summary}
Table~\ref{tab:summ} provides a comprehensive summary of the findings from our survey on existing LLM-based event log analysis techniques. The table provides taxonomy of the findings into various aspects such as anomaly detection, fault monitoring, log parsing, root cause analysis, SIEM, system health monitoring, and threat detection. Each category simply highlights the key methods and approaches discussed in the literature.

\section{Use of LLM in other applications}
To begin,~\cite{wu2024surveylargelanguagemodels} aims to evaluate how well LLMs can correctly recommend content to users based on their behaviour. The methodology was to look at previous publications and see how they performed based on using different techniques and factors. They split LLMs into two groups: discriminative (models suited towards understanding and analysing) and generative (models suited to NLP tasks). The paper highlights that both groups have their advantages and disadvantages and the choice of which group to choose from will depend on the use-case. Discriminative models are shown to achieve better in situations which require them to analyse information, in this case, user preferences, and then provide relevant recommendations, for example, if the user likes science fiction movies then suggest content within this genre. Generative models are shown to be better suited to providing more information due to their speciality in NLP tasks and they can learn in a context which allows them to perform better with zero-shot and few-shot learning compared to discriminative models. A limitation is that this paper covered the advantages and disadvantages of the two groups but did not conduct testing to show this with statistics. 

A different topic is investigated by~\cite{xiao2024cellagent}, who analyse cells. They use three different GPT-4 agents to perform the tasks of: planning (organising how to perform data analysis), executing (the task of data analysis) and analysing (using the model to evaluate the previous step instead of human evaluation). In addition, they store previous data, to be used by the executor agent, intending to increase accuracy. Various event log analysis implementations using LLMs require humans to evaluate the responses, this study could serve as an inspiration for future work by implementing a LLM agent which would evaluate to address this implementation. A similar approach, although with one agent, is highlighted by~\cite{dolphin2024extracting} who use LLMs to produce ``structured insights for financial news''. Their method begins by collecting various relevant articles and parsing them, then implementing prompt engineering to extract relevant information from the article and finally, acronyms (``tickers'') are validated by comparing their string values to that of the full string (string similarity approach). This method was compared to its predecessors and results showed that this was an improvement of 90\%. A limitation of this method is that it is not clear how articles are updated on an hourly basis, as a suggestion, this could be achieved using RAG. In regards to applying this research to event log analysis, future work could investigate the use of RAG, to update the model's knowledge base (with new event logs) and the concept of string similarity can be used to allow the model to continue to understand future event logs after their format has changed slightly. An example is if a certain part of a log could be abbreviated, researchers can have its abbreviation matched to the full version and if this change is present in logs retrieved by RAG, then the model will still understand its meaning. 

Proceeding, we look at various use cases where social media posts have been used to perform analysis in various fields.~\cite{rao2024quallm} propose a general method to evaluate online posts, to establish people's views on different subjects, using LLMs. They propose an approach of using in-context learning where prompts were given in groups (to avoid large costs as a result of abundant API calls). The researchers evaluated the model and they found that it had a factuality score of 0.55, completeness score of 0.78 and accuracy of 0.74. Limitations of the method are the time it would take to teach the model, as well as the authors highlighting that outputs are not fixed, which could result, in outputs, which differ at times. Future work based on this paper can look into using other methods such as fine-tuning and RAG which could improve scores and reduce the time taken to teach the model. A more specific approach is taken by~\cite{radwan2024predictive}, who utilise a combination of LLMs and machine learning to identify if social media posts are suggestive of mental health disorders. The methodology was for the LLMs (GPT-3, BERT and mBERT) to receive the posts and then convert them into embeddings, subsequently, these embeddings were inputted into various machine learning models to classify them as indicating various mental health disorders or normal behaviour. Each model was tested with a variety of different machine-learning techniques to establish which combination was the best for the model. The results showed that GPT-3 in combination with SVM performed the best, followed closely by BERT with LSTM, and mBERT with CNN-BiLSTM performed the worst. A limitation of this study is that the datasets used for training and testing differed between models. Future work based on this paper can look into the use of other techniques such as fine-tuning and RAG to improve performance and accuracy. Finally,~\cite{jaradat2024multitask} research classifying social media posts as indicating road traffic incidents. The methodology was to gather and pre-process the data, then use GPT-3.5 to automatically label the data. This data was used to fine-tune GPT-2 and it was compared to GPT-4 with zero-shot learning and the machine learning model XGBoost. The results show that the fine-tuned GPT-2 model achieved the highest accuracy of 85\%, followed by XGBoost which had an accuracy of 83.5\% and finally GPT-4 with zero-shot learning had 64\%. These results showed that fine-tuning significantly improved performance compared to the base model, even though an early iteration of the model was used. A limitation of this study is that the data was specifically taken from Twitter (known as X) which could mean it is biased, as it could be restricted due to certain demographics using the platform whereas others may not. Future work, specific to this study, could use a larger range of social media platforms to compile the dataset which would reduce the risk of biases. 

To conclude,~\cite{xie2024waitgpt} aims to make performing data analysis tasks easier, due to LLMs, people who are not from a programming background are now able to complete data analysis attacks. Non-coding personnel find it difficult, at times, to understand the output of the LLM and decipher when it makes errors. The researchers propose WaitGPT which produces diagrams whilst the LLM is generating code. A variety of participants were selected to test WaitGPT and found in general that non-experts viewed it positively and experts (people from a computer science background) still viewed it highly but suggested including features which explain how the code is being generated. Future work based upon this article can look at ``introducing a 'stop' mechanism'' which would allow users to stop the generation, if they observe an error, in addition to, introducing features which can explain, with more detail, the reasoning behind code generation that is generated. In addition,~\cite{pagnoni2024bytelatenttransformerpatches} introduce a new transformer architecture, known as Byte Latent Transformer (BLT), to improve upon the previous token-based architecture. Their architecture involves taking the data and converting them into patches. The size of these patches is based on their complexity, the more complex the data is, the larger the size of the patches. The advantages compared to tokenisation are that models using this architecture have a higher chance of completing complex tasks accurately and successfully and that if parts of the data are less complex then it will be a lower size resulting in more efficient models that consume fewer resources. The results show that the use of this architecture provides significant improvements when compared to tokenisation. Future work can look at using BLT for event log analysis tasks to address limitations such as large resource consumption and difficulty in understanding complex log data.

\section{Current Challenges and Potential Solutions}
\subsubsection{Improving Output Accuracy} \hfill\\

\textbf{\textit{Challenge}} -- Due to some research suggesting that there are a large amount of false positive outputs and current solutions, for the vast majority, requiring human supervision, improving output accuracy is a vital avenue for future research. An example of false positives in event logs using LLMs is if a log contains content which is different to that which the LLM was trained or fine-tuned upon, then it will class it as an anomaly regardless of whether it is or not. The main objective of using LLMs, to perform event log analysis, is to improve efficiency and reduce human involvement and the presence of these limitations in current research is a barrier to achieving this objective.

\textbf{\textit{Potential Solutions}} -- A potential solution to address these issues is to initially use human supervision to evaluate if the model produces correct responses or those which they would approve of. Then use reinforcement learning, as proposed by~\cite{sun2023reinforcementlearningerallms} to learn, based on this feedback, and produce outputs that are suitable and accurate. This will reduce the risk of false positives, as well as, provide an alternative to manually reviewing all outputs of a model. 

In addition to this, research can be conducted to find measures, that address the reasons why these false positive outputs occur, directly. Examples of this could be, that the models match the input logs against training data rather than using reasoning or the training data does not contain a large variety of log types, resulting in the LLM struggling to comprehend more complex logs. Both of these issues will be discussed in the following sections. 

\subsubsection{Improving Quality and Availability of Datasets}\hfill\\

\textbf{\textit{Challenge}} -- Event log data is sensitive, as a result of this,  users will be hesitant to provide this data, due to it containing information regarding activities that they partake in. In addition to this, datasets, which are currently available, are restricted to certain companies, who may not use certain aspects of the operating system and will not use all applications and account for all forms of network traffic. This means that when LLMs are trained they may not understand logs which deviate and differ from those that were used for training or fine-tuning, and as a result of this, they may not be entirely accurate. 

\textbf{\textit{Potential Solutions}} -- A potential solution to this limitation is anonymising event log data, before training or fine-tuning. An example of this is~\cite{portillo2019towards} who check if certain keywords are present within the log. Then they, apply various techniques, such as hashing and tokenization, to anonymize the data. This ensures that users cannot be identified by the data that the LLM was trained upon, resulting in a higher chance of users providing their event log data. In addition, a consideration, when compiling event log datasets, is to obtain logs from various companies, which differ in how they use systems, resulting in more accurate event log analysis with LLMs, as they were trained on logs which cover a variety of different activities. 

\subsubsection{LLMs Understanding and Use of Reasoning}\hfill\\

\textbf{\textit{Challenge}} -- The main aim of deciding to use an LLM, in the case of our focus event log analysis, is to leverage their ability to, after training, perform tasks based on their intelligence. However,~\cite{cometti2024real} suggests that there are a large amount of false positives in event log analysis using LLMs, due to matching input data, against that which was used for training and fine-tuning, or matching against keywords that were in the training dataset. In addition, research conducted by~\cite{mirzadeh2024gsm} also supports this observation as they mention that LLMs were evaluated on various examinations and performed well, the researchers performed the same examinations, with changing values, not the questions themselves and found the results were significantly lower.  

\textbf{\textit{Potential Solutions}} -- The research that brings attention to this limitation, suggests that it occurs due to, pattern matching and model understanding based on the proximity of vectors rather than on knowledge. This explains why LLMs struggle with data that differs from training/fine-tuning datasets, as the vector values will differ and result in models not understanding that they are related. This potentially could be addressed with significant research into developing an alternative architecture and method where models complete tasks, based on knowledge rather than relying on vector proximity. 

\subsubsection{Explaining Output}\hfill\\

\textbf{\textit{Challenge}} -- Event logs are complex and require expertise to understand them, this means that professionals are needed which has costs associated with it and if they are unavailable, then there is a security risk. In addition, there are large amounts of logs produced, which will take a considerable amount of time to analyse and will prevent them from performing other tasks and result in operations not being conducted efficiently. Another concern is that professionals experience fatigue and the possibility of potentially missing logs which require attention. To solve this issue research has been done that explains the format of event logs to people who are not experts, but, as mentioned in the literature review, this explanation provided is at times, complex and difficult for users to comprehend. As a result of this, these models need to produce an output that can be understood by those who lack expertise, without compromising important details, to ensure that if a log requires attention then this is properly communicated. 

\textbf{\textit{Potential Solutions}} -- A potential solution to this challenge is to gather data from security experts containing an explanation of various event logs and to evaluate whether people who are not experts can understand them. When an acceptable amount of data has been collected, it can be used, with in-context learning~\cite{liu2021makes}, to teach the model how to explain its outputs in a way that is easy to understand and, at the same time, contains sufficient information. In this situation, in-context learning would be a better alternative to fine-tuning, as the data gathered will not be enough to perform the task, and doing so will consume a large amount of resources unnecessarily. 

\subsubsection{Addressing Redundancy}\hfill\\

\textbf{\textit{Challenge}} -- Throughout the literature review, we have investigated research which mainly relied upon training or fine-tuning LLMs, using event log datasets. A common and frequent limitation was that LLMs became redundant over time. This is due to the contents of logs changing, as when new software or new features are introduced to existing software, the LLM would not recognise the new logs as it was not trained upon these data. In addition, log formats for the same event may differ, as abbreviations or the full format of words within logs would be used during training, resulting in models not understanding logs in the other format.

\textbf{\textit{Potential Solutions}} -- A solution to address new content which could be found within logs, is to use RAG~\cite{zhao2024retrieval}. The use of RAG will allow LLMs to collect new logs and use this to perform various event log analysis tasks, with increased accuracy. However, compared to other solutions, this is a more complex task due to the limited number of event log datasets available. Research can investigate and develop a method to automatically anonymize event logs which in turn would result in a higher likelihood of companies providing their data, which can be retrieved using RAG.

To address differing formats of logs, we suggest string similarity, as implemented in a different field by~\cite{dolphin2024extracting}, as a potential solution to address this limitation. Both abbreviations and the full version of words contained within logs will be compiled, and when models do not understand a word, it will check to see if there is a similar string and find the correct format. In theory, this should address the limitations mentioned throughout the literature we have gathered as it allows the model to understand all event logs even if they differ from the data that it was trained or fine-tuned upon.

\subsubsection{Analysing Logs to Establish System Health}\hfill\\

\textbf{\textit{Challenge}} -- Currently, no research has looked at the use of LLMs and event logs to establish if a system is healthy. The health of a system could be considered to be the extent to which a system is running compared to its optimal performance, this is shown to be an important metric to determine system health as~\cite{khan2018review} mention it frequently within their article to assess how healthy systems are. If something is found to be causing a decrease in performance (indicating poor system health), it can be identified by the LLM and actions can be taken to resolve this and restore the system to being in a state of good health. 

\textbf{\textit{Potential Solutions}} -- Event logs are suitable to be used as an indicator of system health, as they show everything that occurs in a system. Patterns can be identified; for example, if X series of events occur, there is likely a threat to system health and actions need to be performed to remedy and stop this threat from occurring. The LLM can then either perform these actions itself, if it has the permission to do so or suggest advice on how to do this. Both of these suggestions will allow the LLM to successfully diagnose threats to system health and reduce the workload on security professionals and administrators, as it will either resolve or provide information on how to resolve tasks reducing the need for this to be completed manually.

\subsubsection{Inputting Data when Generated}\hfill\\

\textbf{\textit{Challenge}} -- The main objective of using LLMs to perform event log analysis is to improve efficiency and reduce the workload of security personnel who spend considerable time, manually checking event logs. As event log data are produced on a large scale manually inputting into LLMs will still take a large amount of time and also comes with the risk of missing out data, which compromises the security of the systems. As a result of this, it is important for future research to address the challenge and allow event log data to be automatically entered into LLMs. 

\textbf{\textit{Potential Solutions}} -- A thorough literature search shows that, at the time of writing this article, no research has been conducted on the automatic input of data into an LLM. The results were restricted to automatic gathering of data as it is generated for training and fine-tuning. As a result of this, a potential solution, based on~\cite{coutinho2024role}, could be to develop a piece of AI-based software that collects event logs as they are generated and inputs them into the LLM.

\subsubsection{LLM Security}\hfill\\

\textbf{\textit{Challenge}} -- As event logs are sensitive data, it is essential to ensure that anything which uses the data is secure, in the case of our paper LLMs. We have frequently discussed, that open-source LLMs, such as ChatGPT, pose a risk due to the data being sent to their servers, but have mentioned that other research resolves this limitation as they use close-source models. A remaining security concern is that there is a risk of data leaks due to prompts received from malicious actors~\cite{wang2024pandoraswhiteboxprecisetraining}.

\textbf{\textit{Potential Solutions}} -- A solution to this challenge is to use a firewall such as GPTWall proposed by~\cite{li2024governing}. This provides developers with the ability to set limitations to what content would be produced within outputs. In regards to our context, this can be set to not show user-identifiable information and other content within logs themselves, this will ensure that outputs only contain necessary information.

\subsubsection{Covering A Range of Log Types}\hfill\\

\textbf{\textit{Challenge}} -- As seen throughout this paper, current research focuses on specific forms of event log analysis rather than multiple forms, for example, anomaly detection and others on root cause analysis. This is mainly because the research field is due to a recent development and researchers decided to improve performance on tasks individually ensuring that they were able to reach their potential. To continue to innovate, future research can develop LLMs that can perform multiple event log analysis tasks.

\textbf{\textit{Potential Solutions}} -- The solution to this challenge is to develop an LLM model that can perform two or more different event log analysis tasks. This can begin from being two different tasks and then the number can be increased with further development, to ensure that both tasks work correctly. The main difficulty in implementing multiple tasks is all research articles reviewed use either training, fine-tuning or in-context learning to teach the models how to perform the various tasks. In the case of multiple tasks, the model could encounter difficulty deciding which instructions are relevant. This could be resolved by asking users which task they want to complete and ensuring that the model adheres to the relevant training bounds until they are prompted otherwise.

\section{Conclusion and Future work}
This paper provided an in-depth overview of the progress of using LLMs for event log analysis. Early work showed that various models were suitable for event log analysis and their performance was comparable to research at the time. Following this, researchers highlighted that proprietary models such as GPT and Claude were not suitable due to compromising security, compared to open-source models, which do not require data to be sent or stored on the developer's servers. Different factors affecting LLMs, such as in-context learning, fine-tuning and RAG, were then studied. Research showed that in-context learning is viable but performs significantly worse, compared to fine-tuning, which produced significant results. Most research achieved F1 scores ranging between 0.9-1 (compared to previous research which was for the most part below 0.9) and the vast majority where in the upper bound of this limit. Research into RAG is in its early stages but beneficial because it addresses the common limitation of the format an content changing over time.  In addition, it also showed that the use of RAG necessitates combining it with fine-tuning to achieve successful results.

Research into this area is still in very early stages and many limitations can be addressed to contribute to advancing knowledge. A major research area could address the limitations of redundancy and sourcing event logs to be used as datasets for training, fine-tuning, and retrieval with the implementation of RAG. If a source of event logs manifests, that is secure and contains many logs which update over time, it can be used for these purposes and address the two limitations. In addition, an important advancement would be the automatic input of event logs as they are generated. This eliminates the need for this to be done manually, addressing the intended purpose of this field of study which is to improve the efficiency of event log analysis.

\bibliographystyle{ACM-Reference-Format}
\bibliography{bib}

\end{document}